\documentclass[12 pt,letterpaper]{article}
\usepackage{named} %
\usepackage{graphicx} %
\title{Informal Concepts in Machines\thanks{This work is licensed under the Creative Commons Attribution-No Derivative Works 3.0 Unported License (see http://creativecommons.org/licenses/by-nd/3.0/).}}
\author{Kurt Ammon\footnote{Correspondence to paper at cstruct period org. Comments
are welcome.}} %
\date{} %

\newcommand{\bc}{\begin{center}}
\newcommand{\ec}{\end{center}}
\newcommand{\bci}{\begin{center} \begin{minipage}{11.6cm} }
\newcommand{\eci}{\end{minipage} \end{center} }
\newcommand{\be}{\begin{enumerate}}
\newcommand{\ee}{\end{enumerate}}
\newcommand{\beq}{\begin{equation}}
\newcommand{\eeq}{\end{equation}}

\newtheorem{dfn}{Definition}
\newcommand{\bdf}{\begin{dfn}\rm}  
\newcommand{\edf}{\end{dfn}}  
\newtheorem{thm}{Theorem}
\newcommand{\bth}{\begin{thm}\rm}  
\newcommand{\eth}{\end{thm}}  
\newtheorem{hyp}{Hypothesis}
\newcommand{\bhp}{\begin{hyp}\rm}  
\newcommand{\ehp}{\end{hyp}}  
\begin{document} %
\maketitle
\begin{abstract}
{\it This paper constructively 
proves the existence of an 
effective procedure %
generating a computable (total)
function %
that is not contained in any given effectively enumerable set of such functions. %
The proof implies the existence of machines that process informal concepts such as computable (total) 
functions beyond the limits of any given Turing machine or formal system, that is,  
these machines can,  in a certain sense, ``compute" function values beyond these limits.
We call these machines creative.  We argue that any ``intelligent" machine should be capable of processing informal concepts such as computable (total) functions, that is, it should be creative. 
Finally, we introduce hypotheses on creative machines which were developed on the basis of theoretical investigations and experiments with computer programs. 
 The hypotheses 
say that machine intelligence is the execution of a self-developing
procedure starting from any universal programming language and any input.
}
\end{abstract}

\section{Introduction}  
                     
\noindent
Hilbert's program aimed %
to reduce mathematics to a formal system in order to avoid inconsistencies in mathematics. 
In particular,
Hilbert's Entscheidungsproblem (decision problem) aimed to find ``a procedure that allows one to decide on
the validity, respectively satisfiability, of a given logical expression by a finite
number of operations".\footnote{Hilbert and Ackermann 
\shortcite[p.\ 73]{HilbertAckermann28}: 
ein Verfahren ..., %
das bei einem vorgelegten logischen Ausdruck durch endlich viele Operationen die Entscheidung \"uber die 
Allgemeing\"ultigkeit bzw.\ Erf\"ullbarkeit erlaubt. 
}
In order to prove that Hilbert's Entscheidungsproblem is unsolvable,
Turing \shortcite{Turing36} introduced his ``computing machines" which are a formalization of a
 {\em procedure} in Hilbert's sense.
G\"odel \shortcite[p.\ 72]{Goedel65} writes: 
\bci
Turing's work gives an analysis of the concept of ``mechanical procedure" (alias ``algorithm" ...) ... This concept is shown to be equivalent with that of a ``Turing machine". A formal system can simply be defined to be any mechanical procedure for producing formulas, called provable formulas.
\eci

Hopcroft and Ullman \shortcite[p. 147]{HopcroftUllman79} write that ``the Turing machine is equivalent in computing power to the digital computer as we know it today". They implicitly assume that the computer is {\em used} 
for executing a  {\em given procedure} or program that was developed manually.
Turing \shortcite{Turing47} asks whether a computer can be {\em used\/} in another way:
\bci
It has been said that computing machines can only 
carry out the processes that they are instructed to do. 
... %
Up till the present machines have only be been
used in this way. But is it necessary that they should always
be used in such a manner?
\eci
Turing \shortcite{Turing48} 
discusses the development %
 of intelligence in man and in machines:
\bci
If the untrained infant's mind is to become an intelligent one, it must acquire both
discipline and initiative. So far we have been considering only discipline. 
To convert a brain or machine into a universal machine is the extremest form of discipline. 
But
discipline is certainly not enough in itself to produce intelligence. That which is
required in addition we call initiative. ... %
Our task is to discover the nature of this residue as it occurs in man, and to try and copy
it in machines.
\eci
We investigate ``the nature of 
this residue" called ``initiative"
(see Sieg, \citeyear[Section 5, Final remarks]{Sieg94}).
Section \ref{ft} \mbox{} proves
the existence of 
an effective procedure %
generating a  computable (total)
function %
that is not contained in any given effective enumeration of such functions. 
This procedure %
can be regarded as a %
 bridge %
to
the informal concept of computable (total) functions, that is, to
Turing's 
uncomputable
residue. 
On the basis of this 
procedure %
Section \ref{cm} defines creative machines which can, 
in a certain sense, ``compute"
function values beyond the limits of any given Turing machine.
 Section \ref{hyp} introduces hypotheses on creative machines which %
say that Turing's uncomputable residue is the execution of a self-developing
procedure starting from any universal programming language and any input. This process,
which produces formally irreducible experience, can be regarded as a new use 
of computers.
The remaining sections discuss our proof, creative machines and related work.

\section{First Theorems}\label{ft} 

In this and the following sections 
we simply write {\em computable function} for an effectively computable {\em total} function
of natural numbers which is defined for {\em all} natural numbers.
\begin{thm}\label{cen}
There is an effective procedure %
generating a  computable 
function %
that is not contained in any given effective enumeration 
of such functions.
\end{thm}
{\bf Proof}. 
Let $f_1, f_2, %
...$ %
be an effective enumeration of  computable 
functions. %
We define a new function 
$g$ %
 by 
\beq
 g(n)=f_n(n)+1 \label{g=fn+1}
\eeq 
for all natural numbers $n$. 
Obviously, $g(n)$ is defined for all natural numbers $n$ because $f_i(n)$
is defined for all natural numbers $i$ and all natural numbers $n$. Furthermore, $g$ is computable
because $f_1, f_2,%
...$ %
is an effective enumeration
of  computable functions
according to our original 
assumption. 
The expression
 $f_n(n)+1$ \label{fn+1}
in the definition (\ref{g=fn+1}) of $g$ can be regarded as a functional pseudocode,
that is, as a computer program, say $R$, that computes the function $g$ for all natural numbers $n$.
There is
an effective procedure %
that generates %
the program $R$.
This procedure can be represented as a computer
program
whose input is the effective enumeration  $f_1, f_2, ...$, that is,
a program, say $E$, generating the functions $f_1, f_2, ...$,
and whose output is the program $R$. %
In order to generate $g(n)$ from any natural number $n$, 
the program $R$ thus generates the function
$f_n$ by applying $E$ to $n$
and then adds $1$ to the result of applying $f_n$ to $n$.
Because of definition (\ref{g=fn+1}), 
$g(n)$ is different from $f_n(n)$ for all natural numbers $n$.
This implies that
the  computable function $g$ is different from  
all functions $f_n$, where $n$ is any natural number.
Therefore, there is an effective procedure %
generating a  computable 
function $g$ %
that is not contained in any given effective enumeration
of such functions.

\begin{thm}\label{cfs} 
There is an effective procedure %
generating a  computable 
function %
that is not contained
in any given formal system with a predicate for such functions. 
\end{thm}
{\bf Proof}. Let $S$ be a formal system with a predicate $Q$ for  computable functions.
Because a formal system can be regarded as a mechanical procedure or algorithm producing provable formulas
(G\"odel \citeyear[p.\ 72]{Goedel65}),
the formal system $S$ produces an effective enumeration of provable formulas $Q(f_1), Q(f_2), 
 ...$ 
that contains 
an effective enumeration of 
all  computable functions $f_1, f_2, %
...$ %
in 
$S$.
According to Theorem \ref{cen} there is an effective procedure %
generating a  
 computable function %
that not contained 
in the effective enumeration of 
all  computable functions $f_1, f_2, %
...$ %
in $S$.

\medskip %
Theorem \ref{cfs} implies that the concept of computable functions
is informal in the sense that
the effective procedure in the theorem
generates a computable function
beyond the limits of any given formal system.

\section{Creative Machines}\label{cm} 

We apply Theorem \ref{cen} to a hypothetical machine $C$
capable of processing an effective procedure %
that exists according to Theorem \ref{cen}.
There are no assumptions on the internal structure of $C$. %
In particular, it is not assumed that 
$C$ %
is a Turing machine in 
any sense. 

\begin{thm}\label{cmTm}
Let $C$ be a machine capable of processing an effective procedure $P$ %
that exists according to Theorem \ref{cen}.
Then, there is no Turing machine generating all  computable 
functions that $C$ can generate by means of $P$.\footnote{The theorems and proofs will be described more precisely in a separate paper.}
\end{thm}
{\bf Proof}. Let $C$ be a machine capable of processing an effective procedure $P$ %
that exists according to Theorem \ref{cen} and 
let $T$ be any Turing machine generating 
any enumeration
$f_1, f_2, %
...$ 
of  computable functions. %
The enumeration $f_1, f_2, %
...$ is effective because it is generated by a Turing machine.
The machine $C$ can generate a function $g$ by applying 
the effective procedure
$P$ to this effective enumeration.
According to Theorem \ref{cen} the function $g$ is a  computable function %
that
is not contained in the enumeration  $f_1, f_2, %
...$ of computable functions.
Therefore, there is no Turing machine $T$ generating all  computable 
functions that $C$ can generate by means of $P$.

\begin{thm}\label{cmfs}
Let $C$ be a machine capable of processing an effective procedure $P$  
that exists according to Theorem  \ref{cen}.
Then, there is no formal system containing all  
computable functions that $C$ can generate by means of $P$.
\end{thm}
{\bf Proof}. Let $C$ be a machine capable of processing an effective procedure $P$ 
that exists according to  Theorem \ref{cen} and let $S$ be any formal system 
with a predicate $Q$ for computable (total) functions. %
Because a formal system can be regarded as a mechanical procedure or algorithm producing provable formulas
(see G\"odel \citeyear[p.\ 72]{Goedel65}),
the formal system $S$ produces an effective enumeration of provable formulas $Q(f_1), Q(f_2),  
...$ 
that contains all  computable functions $f_1, f_2, %
...$ %
in the formal system $S$.
The machine $C$ can generate a function $g$ by applying the effective procedure $P$ 
to the effective enumeration
$f_1, f_2, %
...$ of functions.
According to Theorem \ref{cen} the function $g$ is a  computable function %
that
is not contained in the enumeration $f_1, f_2, %
...$ of functions.
Therefore, there is no formal system $S$ containing all  computable 
functions that $C$ can generate by means of $P$.
\medskip %

Because all programs of a %
programming language are finite sequences of a fixed finite number
of symbols, they
 can be effectively enumerated, for example, in ascending length. This implies that there is an effective enumeration 
of all computable partial functions which need not be defined 
for all natural numbers. %
                       Thus, Theorem  \ref{cen}
implies that the function deciding whether or not a 
computable partial function %
is a total function is uncomputable, that is, not computable.
If this function were computable, its application to an effective enumeration
of all computable partial functions would yield an effective enumeration of all  
computable (total) functions. %
This contradicts 
Theorem  \ref{cen}.
For these reasons,
the function deciding whether a computable partial function is a total function is uncomputable. %
Thus, the function $g$ in the proofs of Theorems \ref{cmTm} and  \ref{cmfs},
which is generated by the hypothetical machine $C$,
is a value of this uncomputable function that is not contained in the given
Turing machine  $T$  and the given formal system $S$, respectively.
Therefore,
Theorems \ref{cmTm} and  \ref{cmfs} 
imply %
that 
the hypothetical machine $C$
can compute
values of this uncomputable function beyond the limits of any given Turing machine or formal system.
This suggests the following definition for a new kind of machines.
\bdf\label{dcm}
A machine is called {\it creative} if it is capable of evaluating functions, that is, 
determining function values beyond the limits of any given Turing machine or formal system.
\edf
In view of  Theorem \ref{cmTm} %
creative machines can process the informal concept of  computable functions beyond the 
limits of any given Turing machine. %
In particular, the effective procedure $P$ in the theorem %
can be regarded as a bridge
to this informal concept, that is, to
Turing's uncomputable ``residue". 
The next section 
introduces hypotheses on the development of creative machines and their mode of operation.
The hypotheses provide more information on the nature of Turing's ``residue".

\section{Hypotheses}\label{hyp} 

Let $P$ be a computer program whose properties are not known, that is, 
we have no or only partial knowledge about $P$. 
We can apply $P$ to an input, for example, the natural number $1$, which may produce  
the natural number $2$. Thus, the execution of $P$ produces knowledge  which can be represented in function notation
by
\beq\label{p12}
P(1)=2.
\eeq
We may apply the program $P$ to another input,
 for example, the natural number $2$, which may produce  
the natural number $3$, that is, the knowledge
\beq\label{p23}
P(2)=3. 
\eeq
From (\ref{p12}) and (\ref{p23}), we may assume by fallible inductive reasoning that
\beq\label{pn+1}
P(n)=n+1 
\eeq
holds for all natural numbers $n$.
This simple example illustrates how 
knowledge about a program can be produced by the execution of the program and fallible inductive reasoning.

In order to develop a computer program a programmer usually applies 
all available knowledge. When the program is written, 
he tests it to verify whether it has the desired 
properties, that is, he has only partial knowledge about the program.
In his tests he executes the program which produces further knowledge,
for example, whether it generates an output from any input.
Finally, he concludes on the basis of the tests and all available knowledge
that the program has the desired properties.
This conclusion can be regarded as fallible inductive reasoning.
Thus, program development usually involves the application of all available knowledge,
the execution of the program in tests to produce further knowledge, 
and fallible inductive reasoning to conclude whether the program has the desired properties.

The expression $n+1$ in (\ref{pn+1}) can be regarded as a program
computing the natural number $n+1$ from any natural number $n$.
It can be constructed from
the elementary knowledge
that $n$ and 1 are natural numbers and $x+y$ is a natural number for any natural
numbers $x$ and $y$.
If we regard the variable $n$ und the constant $1$ in the expression $n+1$
as nullary functions (without arguments) whose values are natural numbers,
the expression $n+1$ in (\ref{pn+1})
is just the composition
of the functions $n$, 1, and $+$ that are contained in %
$n+1$.

The composition of functions can be used as an elementary mechanism for 
the construction of sophisticated programs. For example, 
Ammon \shortcite{Ammon88} describes the automatic development of
a program that proves theorems in mathematics whose complexity represented the state of the art 
in automated theorem proving.\label{hypAmmon}
The program is also constructed on the basis of elementary 
functions that form the components of the program such as the ``{\em left-side}"
$x$ of an equation $x=y$ (see Ammon \citeyear[p.\ 559, Table 2]{Ammon88}).
Starting ``from scratch" compositions of the elementary functions are used to construct 
"a sequence of more and more powerful partial methods [programs] each
of which forms the basis for the construction of its successor
until a complete method [program] is generated" \cite[p.\ 558, Abstract]{Ammon88}.
The elementary functions themselves which form the components of the program,
for example, the ``{\em left-side}"
$x$ of an equation $x=y$,
 can be constructed on the basis of the elementary instructions or functions of a programming
language.

The preceding considerations formed a starting point for the following hypotheses about 
creative machines.
The next sections provide further arguments.
In the hypotheses, a programming language 
is meant to include elementary knowledge about the language, for example,
the domains and ranges of its elementary instructions or functions
which can be used to form compositions of the instructions or functions.
Roughly speaking, the first hypothesis states that
the knowledge in a creative machine is developed in a formally irreducible empirical 
process from a programming language.

\bhp {\bf (Experience)} \label{hex} 
A creative machine is the execution of a self-developing
procedure including knowledge
which starts from any universal programming language and any input. This process
produces formally irreducible experience, that is, the self-developing
procedure cannot be reduced to 
a formal system but the language and the input from which it starts.
The development of the procedure can be illustrated by the formula
\beq
L+P_t+E \rightarrow P_{t+1},
\eeq
where $L$ is the universal programming language including knowledge about $L$ itself, 
$P_t$ is the procedure at time $t = 0, 1, 2, ...$,
$P_0$ is empty, %
and $E$ stands for experience.
\ehp
The second hypothesis refers to informal concepts such as the  
computable functions in Theorems \ref{cmTm} and  \ref{cmfs}.
\bhp {\bf (Structure)}\label{hst} 
The knowledge in a creative machine according to Hypothesis \ref{hex}
contains informal concepts that are extensible beyond the limits of any given
Turing machine or formal system.
\ehp
The third hypothesis
deals with inductive reasoning, that is, the
construction of knowledge in a creative system.
\bhp {\bf (Induction)}\label{hin} 
In practice the construction of knowledge in a creative machine is achieved by
informal inductive reasoning that is 
based on 
all available knowledge including informal concepts 
according to Hypothesis \ref{hst}.
\ehp
The last two hypotheses say
 that a creative system can 
revise all its knowledge
and
construct any knowledge.
\bhp {\bf (Revision)}\label{hre} 
A creative machine may revise all its knowledge
but a universal programming language from which it starts
according to Hypothesis \ref{hex}. In principle, all knowledge
is fallible and correctable by the machine.
\ehp
\bhp {\bf (Generality)}  \label{hge} 
A creative machine can in principle 
construct and verify any knowledge
including all knowledge about itself
as far as the knowledge can be constructed and verified.
\ehp

\section{Discussion}\label{dis} 

We discuss creative machines, in particular, in view of the Church-Turing thesis.
The next section compares related work, for example, the Turing machine concept.

In order to apply an effective procedure $P$ %
according to 
Theorem \ref{cen} in Section \ref{ft}, a creative machine $C$ according  
to Definition \ref{dcm}  in Section \ref{cm} %
needs an effective enumeration of  computable functions, say $E_1$.
A simple example is an enumeration of functions $f_1$, $f_2$, %
... defined 
by $f_i(n)=i$ for all natural numbers $i$ and natural numbers $n$.
By applying the effective procedure %
$P$ to $E_1$ the machine $C$ 
can generate a  computable function, say $g_1$, which is
not contained in $E_1$ according to 
Theorem \ref{cen}. The function $g_1$ can be added to $E_1$ which yields an effective enumeration 
$g_1$, $f_1$, $f_2$, %
..., say $E_2$. 
Thus, 
starting from any effective enumeration $E_1$
of  computable functions a creative  machine $C$ can, by repeatedly applying $P$, 
generate a sequence  $E_1$, $E_2$, %
... of effective enumerations
of  computable functions
where $E_n$ has the form %
 $g_{n-1}$, ..., $g_1$, $f_1$, $f_2$, %
... for
any natural number $n$ greater than 1.
 In particular, each $E_n$ %
contains another  computable function $g_{n-1}$ that 
in not contained in any preceding $E_i$, where $i$ is a natural number less than $n$.

An effective enumeration $E_1$ of  computable functions can be provided by a
Turing machine, say $T_1$ generating the functions in $E_1$ or
by a formal system, say $F_1$, with a predicate for the computable functions in $E_1$.
By applying an effective procedure $P$ %
according to 
Theorem \ref{cen} a creative machine $C$ 
can generate a  computable function, say $g_1$.
Theorems \ref{cen} and  \ref{cfs} in Section \ref{ft} imply that
the function $g_1$ is neither generated by 
the Turing machine $T_1$ nor contained in the formal system $F_1$.
The incorporation of $g_1$ into the Turing machine $T_1$ and the formal system $F_1$
yields a more powerful Turing machine $T_2$
and a more powerful formal system $F_2$. %
Thus, 
a creative  machine $C$ can, by repeatedly applying $P$, 
generate a sequence of more and more powerful Turing machines  $T_1$, $T_2$, %
 ...
and a sequence  of more and more powerful formal systems $F_1$, $F_2$, %
 ..., 
respectively.

Up to any point in time a creative system $C$ can only ``know"
a finite description of  computable functions which can 
be represented by a Turing machine, say $T_1$, generating the functions
or in a formal system, say $F_1$, with a predicate for the functions.
As described above the creative machine $C$ can, by repeatedly applying 
an effective procedure $P$ according to Theorem \ref{cen} to $T_1$
or $F_1$, generate more and more powerful Turing machines  $T_1$, $T_2$,
...
and more and more powerful formal systems $F_1$, $F_2$, %
..., 
respectively. 
Thus, a creative system $C$ can generate ``knowledge", that is, computable functions
beyond the limits of its own knowledge
in the Turing machine $T_1$ and the formal system $F_1$.

According to our considerations 
preceding Definition \ref{dcm}
in Section \ref{cm},
the propositional function (predicate), say $d$,  deciding whether or not a 
computable partial function of natural numbers is a  total function is uncomputable. %
The preceding paragraphs in this section imply that a creative system  
can determine an unlimited number of values of the function $d$ beyond its ``knowledge", that is, 
it can repeatedly generate  computable functions
beyond the limits of its ``knowledge", which is finitely describable,
and the limits of a given Turing machine or formal system.
The Church-Turing thesis states that every effectively calculable function
is computable by a Turing machine (see Kleene \citeyear[pp. 317-323, 376-381]{Kleene52}). 
A creative machine can determine values of the function $d\/$ beyond its finitely describable 
"knowledge" but it cannot calculate the value of the function $d(x)$
"for each value of $x$ for which it is defined" (see Kleene \citeyear[p.\ 493]{Kleene87}), that is, 
it cannot determine for each computable partial function $x$ whether $x$ is a total function.
Therefore, a creative system can compute values of uncomputable functions 
beyond the limits of given Turing machines or formal systems, but this has no 
bearing on what number-theoretic functions are effectively calculable (see Kleene \citeyear[p.\ 493]{Kleene87}).
Furthermore, the  computable functions
generated by a creative machine
cannot be specified {\em in advance} but must
be generated by repeated applications of an effective procedure according to Theorem \ref{cen}.
This process involves the generation of a sequence of more and more powerful 
effective enumerations of  computable functions, equivalent Turing machines, or equivalent
formal systems each
of which forms the basis for the generation of its successor.
Roughly speaking, this process can be regarded as a self-developing procedure
which must be generated step by step and cannot be reduced to a finite description given in advance
(see Theorems \ref{cmTm} and  \ref{cmfs} in Section \ref{cm}).
Kleene \shortcite[p.\ 493]{Kleene87} requires that an ``effective calculation procedure" or  ``algorithm" is 
"fixed in advance for all calculations" and that it is
"possible to convey a complete finite
description of the effective procedure or algorithm by a finite communication,
in advance of performing computations in accordance with it".
For any  computable (total) function that a creative system can generate it
can give a complete finite description of its generation but there is
no complete finite description for the generation of all  computable functions it
can generate (see Theorems \ref{cmTm} and  \ref{cmfs}). %

Hypothesis \ref{hex} in Section \ref{hyp} states that
the knowledge in a creative machine is developed in a formally irreducible empirical 
process that starts from a programming language and any input.
As described above, the development of knowledge is formally irreducible
because it must be generated step by step and cannot be reduced to a finite description given in advance.
For example, this process can involve the generation of a sequence of more and more powerful 
Turing machines or formal systems each
of which forms the basis for the generation of its successor
(see Theorems \ref{cmTm} and  \ref{cmfs}). %
In a comparable process Ammon \shortcite{Ammon88} generates programs on the basis of elementary 
functions that form the components of the final programs.
Starting from scratch, compositions of the elementary functions are used to
construct a sequence of more and more powerful partial programs each
of which forms the basis for the construction of its successor
until a complete program is generated (see Section \ref{hyp}). %

According to Hypothesis \ref{hst} a creative machine contains informal concepts
such as computable functions. These concepts are processed
by effective procedures such as an effective procedure according to Theorem \ref{cen}.
But according to Theorems \ref{cmTm} and  \ref{cmfs} %
they are extensible beyond the limits of any given Turing machine or formal
system. This means that informal concepts are processed by effective 
(finite) procedures
although there is no complete formal %
(finite) description of these concepts.

According to Hypothesis \ref{hin} 
knowledge is constructed by
informal inductive reasoning that is 
based on 
all available knowledge.
As described in Section \ref{hyp} a
model of such reasoning processes
is the development of computer programs
which are empirically verified in tests and then used in practice.
Such an empirical verification requires only limited resources such as
time.
This means that informal inductive reasoning 
is regarded as more efficient in practice than sophisticated formal procedures.
Because all available knowledge can be used
there is no sophisticated formal method for the construction of knowledge 
that can specified in advance.

The fourth hypothesis implies that a creative machine may start from 
scratch, that is, from any programming language and any input. Thus,
it can revise any knowledge that proves to be false in a specific case.

The last hypothesis says that a creative machine
can
construct and verify any knowledge
including knowledge about itself.
Here, ``any knowledge" includes any knowledge that can be constructed by a human and
is finitely describable.

According to Definition \ref{dcm} in Section \ref{cm}
a creative machine is capable of evaluating functions %
beyond the limits of any given Turing machine or formal system.
This means that it is capable of
evaluating uncomputable functions.
Hypotheses \ref{hin} and \ref{hre} imply that this evaluation %
ordinarily involves empirical
inductive reasoning that is fallible and correctable, that is,
the values $f(x)$ of uncomputable functions $f$ a creative machine assigns to arguments $x$
may be false but can be corrected by the machine.
An example of an uncomputable function is the uncomputable propositional 
function (predicate) that determines whether
a computable partial function is a total function.
The development of computer programs provides a model
for the evaluation of this uncomputable predicate 
because, for example, a program must produce 
an output for any input for which the program is defined.

How can we use our insight into the nature of Turing's residue,
in particular, into the evaluation of uncomputable functions
to implement a creative machine?
Turing \shortcite{Turing48} writes: ``Our task is to discover the nature of 
this residue as it occurs in man, and to try and copy it in machines."
According to our hypotheses we must choose a programming language 
and implement knowledge about this language, for example, the domains and
ranges of the elementary functions or instructions it contains.
In order to reduce the complexity of a first implementation
we should reduce the number and complexity of the informal
concepts to be implemented.
A concrete project might aim to 
prove theorems in a mathematics textbook
on the basis of preceding theorems and proofs (see Ammon \citeyear{Ammon88}).
According to our approach 
no formal system, which eliminates Turing's residue,
should be used but
the ordinary representation of theorems and proofs
should be modeled.
Another possible field of application might aim at an automatic 
development of programs.
An implementation of a creative machine would be general and complete if
it can change all its source code and develop new knowledge from scratch, that is,
from a universal programming language.
Obviously, Ammon \shortcite{Ammon88} %
is far 
from a general and complete system in our sense.

The theorems and hypotheses on creative machines 
have 
epistemological 
implications. 
For example, informal concepts such as  computable functions
are extensible beyond any formal limits.
Because a creative machine can construct and revise any knowledge
starting from any %
programming language and any input
there is only a rather limited general description of its structure and development,
in particular, %
its starting point.
This is comparable with   
Piaget's \shortcite[p. 704]{Piaget70} view:
"Knowledge, then, at its origin, neither
arises from objects nor from the subject,
but from interactions - at first inextricable
- between the subject and those objects."

\section{Related Work}

Ordinarily, the word ``machine" refers to a Turing machine which is a formalization
of the concept of an {\em algorithm} or effective {\em procedure} (see 
Hopcroft and Ullman \citeyear[pp. 146-147]{HopcroftUllman79}).
In contrast, Theorems \ref{cmTm} and  \ref{cmfs} %
contain
no assumptions on the machine $C$
but its capability to process an effective procedure $P$ according to Theorem 
\ref{cen}. 
In particular, the effective procedure $P$ and the machine $C$ are regarded as
different entities.
Thus, Theorems \ref{cmTm} and  \ref{cmfs} can be applied to any machine 
capable of processing the effective procedure $P$.

If one requires that a creative machine can execute any %
Turing machine
it can compute any computable function and evaluate uncomputable functions as %
described in 
this paper.

One might ask whether the machine $C$ in 
Theorem \ref{cmTm} %
can be modeled 
by a Turing machine generating  computable functions, say $T$, 
into which an effective procedure $P$ according to 
Theorem \ref{cen} is incorporated. But according  
to Theorem \ref{cmTm} the machine $C$ 
could apply $P$ to $T$ to generate a  computable function that is not generated by $T$. 
Therefore, the Turing machine $T$ cannot model the machine $C$, that is, it cannot
generate a  computable function beyond the limits of any given Turing machine.

Turing \shortcite[p. 232]{Turing36} writes: ``the motion of the machine ... is {\em completely} determined by the configuration" which comprises a condition from a ``finite number of conditions" and a ``scanned symbol".
The behavior of a creative machine cannot be completely determined in advance
because it contains informal concepts such as  computable functions
which cannot be reduced to a finite description given in advance 
(see Theorems \ref{cmTm} and \ref{cmfs}).
But at any moment and at the most basic level the next ``instruction" is completely 
determined by the present state of a creative machine, which can be regarded  
as a finite string of symbols or binary digits, and the input processed by the creative machine.

Turing \shortcite[pp.\ 249-250]{Turing36} attempts to show that his machines 
can compute ``all numbers which would naturally be regarded as computable"
and supposes that the ``number of states of mind" is finite.
G\"odel \shortcite[p. 306]{Goedel72a}
points out
that ``mental procedures" may ``go beyond mechanical procedures ... 
mind, in its use, is not static, but constantly developing, i.e., that we understand abstract terms more and more precisely as we go on using them ... %
although at each stage the number and precision of the abstract terms at our disposal may be finite, both (and, therefore, also Turing's number of distinguishable states of mind) may converge toward infinity
 in the course of the application of the [mental] procedure". 
The  
``abstract terms" refer to 
abstract concepts such as 
"a certain concept of computable function" %
(G\"odel \citeyear[p. 271]{Goedel72})
and ``the use of abstract terms on the basis of their meaning" (G\"odel \citeyear[p. 72, footnote **]{Goedel65}).
Creative machines can process informal concepts such as  computable functions.
Beyond any given Turing machine or formal system
they can generate an unbounded number of such functions
which can be regarded as a part of the extensible meaning
of this informal concept.

Our paper is no contribution to %
formal logic. Rather, we leave formal logic on the basis of 
Theorems \ref{cmTm} and \ref{cmfs} and our hypotheses on creative 
machines. 
We allow creative machines to apply Cantor's
diagonalization to 
informal concepts such as  computable functions in
Turing machines and formal systems
which are restricted to enumerable subsets of nonenumerable sets that are contained in these concepts.
At any point in time a creative machine can only contain such an enumerable subset %
but it can generate an unbounded number of more and more extended subsets.
A creative machine can simply refute the claim that it is modeled by a 
Turing machine or a formal system by applying an effective procedure 
according to Theorem \ref{cen} to the %
machine or system.

We argue that any ``intelligent" machine should
be capable of processing informal concepts such as
 computable functions 
beyond the limits of any given Turing machine %
or 
formal system, that is, it should be creative.
Formal descriptions 
restrict the generality of a machine,
in particular, 
informal concepts such as  computable functions.

G\"odel's theorem says that 
every sufficiently powerful 
consistent formal number theory contains an undecidable proposition.
Lucas \shortcite{Lucas61} argues %
that mind cannot be modeled by %
 a Turing
machine because he {\em knows} that the %
undecidable
proposition %
is true. %
Putnam points out that Lucas cannot prove the prerequisite of consistency
in G\"odel's 
theorem (see Shapiro \citeyear[pp. 282-284]{Shapiro98}).
The machine $C$ in Theorem \ref{cmTm} %
can process an effective procedure $P$ %
 that produces a computable function $g$
from an effective enumeration of such functions.
This enumeration is comparable with the consistent number theory in 
G\"odel's theorem and the function $g$ with the undecidable proposition.
If a creative machine ``knows" an effective enumeration of
 computable functions it can also ``know" the  computable function $g$ 
which is not contained in the enumeration.
The prerequisite of an effective enumeration of  computable functions
can be  satisfied because there are simple examples of such enumerations
and, up to any point in time, a creative machine $C$ can only ``know"
a finite description of  computable functions which 
can be represented by such an enumeration (see Section \ref{dis}).
Without any %
assumptions on the structure of %
mind 
but its capability to process
 an effective procedure according
to Theorem \ref{cen}
these considerations also apply to humans who can thus extend their knowledge
beyond any formal limits. 
Post's \shortcite[p. 295]{Post44} 
conclusion from G\"odel's theorem
is that ``{\em mathematical thinking is ... %
essentially creative}" (see Shapiro \citeyear[pp. 291-292, ``creative step"]{Shapiro98}).
\section{Conclusion}      
Turing points out that intelligence requires a   ``residue" 
called ``initiative" that is not captured by his machines.
We introduced the concept of a creative machine 
by requiring that it can evaluate uncomputable functions in a certain sense.
It is capable of 
extending informal concepts such as (total) computable functions 
beyond any given formal description which is restricted to incomplete enumerable 
parts of these %
concepts.
We argue that ``intelligent" 
machines %
should %
 be 
capable of processing such %
concepts. %
Hypotheses on creative machines
say that Turing's uncomputable ``residue" is the execution of a self-developing
procedure that starts from any %
programming language and any input. 
This process,
which produces formally irreducible experience, can be regarded as a new use 
of computers.

\bigskip \medskip %
\noindent
{\bf Acknowledgments}. The author wishes to thank colleagues for valuable comments.

\bibliography{Informal}
\bibliographystyle{named} %

\end{document}